\documentclass[letterpaper]{article} 
\usepackage{aaai25}  
\usepackage{times}  
\usepackage{helvet}  
\usepackage{courier}  
\usepackage[hyphens]{url}  
\usepackage{graphicx} 
\usepackage{natbib}  
\usepackage{caption} 
\usepackage{algorithm}
\usepackage{algorithmic}
\usepackage{subcaption}
\usepackage{framed}
\usepackage{amsmath}

\usepackage{newfloat}
\usepackage{listings}
\DeclareCaptionStyle{ruled}{labelfont=normalfont,labelsep=colon,strut=off} 
\floatstyle{ruled}
\newfloat{listing}{tb}{lst}{}
\floatname{listing}{Listing}
\title{Conversation Kernels: A Flexible Mechanism to Learn Relevant Context
for Online Conversation Understanding}
\author {
    Vibhor Agarwal,
    Arjoo Gupta,
    Suparna De,
    Nishanth Sastry
}
\affiliations {
    University of Surrey, Guildford, Surrey, United Kingdom\\
    \{v.agarwal, a.gupta, s.de, n.sastry\}@surrey.ac.uk
}

\newcommand{\Slashdot}{\texttt{slashdot.org} }

\usepackage{xcolor}

\begin{document}

\maketitle

\begin{abstract}
Understanding online conversations has attracted research attention with the growth of social networks and online discussion forums. Content analysis of posts and replies in online conversations is difficult because each individual utterance is usually short and may implicitly refer to other posts within the same conversation. Thus, understanding individual posts requires capturing the conversational context and dependencies between different parts of a conversation tree and then encoding the context dependencies between posts and comments/replies into the language model.

To this end, we propose a general-purpose mechanism to discover appropriate conversational context for various aspects about an online post in a conversation, such as whether it is informative, insightful, interesting or funny. Specifically, we design two families of \textit{Conversation Kernels}, which explore different parts of the neighborhood of a post in the tree representing the conversation and through this, build relevant conversational context that is appropriate for each task being considered. We apply our developed method to conversations crawled from  \texttt{slashdot.org}, which allows users to apply highly different labels to posts, such as `insightful', `funny', etc., and therefore provides an ideal experimental platform to study whether a framework such as Conversation Kernels is general-purpose and flexible enough to be adapted to disparately different conversation understanding tasks.

We perform extensive experiments and find that context-augmented conversation kernels can significantly outperform transformer-based baselines, with absolute improvements in accuracy up to $20\%$ and up to $19\%$ for macro-F1 score. Our evaluations also show that conversation kernels outperform state-of-the-art large language models including GPT-4. We also showcase the generalizability and demonstrate that conversation kernels can be a general-purpose approach that flexibly handles distinctly different conversation understanding tasks in a unified manner.
\end{abstract}

%

\section{Introduction}
Online conversations on social media and discussion forums are an important part of the Web, offering vital emotional support or information-seeking avenues. On many platforms, users can \textit{reply} to posts by other users. Thus, conversations tend to develop as \textit{trees}, where each post (with the exception of the root or original post) has one parent (the post it is replying to), and potentially many children (all the posts replying to it). Such conversations can develop without bound. For example, the BBC News article reporting on former United Kingdom (UK) Prime Minister
Tony Blair’s thoughts on Brexit\footnote{\url{https://www.bbc.co.uk/news/uk-politics-38996179}, last accessed 22 Mar 2025.} had attracted over 10,000 comments. Similarly, there is an average of 42,600 tweets per day exchanged between UK Members of Parliament and their
followers~\cite{agarwal2019tweeting}, emphasizing the  information flow between posts and replies.

Given the scale of such public conversations, there is a need for automated methods for understanding conversations and detecting various kinds of online posts.
Existing efforts for understanding conversations include identifying, for instance, whether a post contains hate speech~\cite{paz2020hate,yin2023annobert,agarwal2023haterephrase,pushkalHateSpeech}, partisanship~\cite{karamshuk2016identifying,agarwal2023graph} or misinformation~\cite{islam2020deep,su2020motivations}.

There is a growing recognition that such conversation understanding tasks require taking into account the wider context of the conversation, not just an individual post in isolation~\cite{perez2023assessing,agarwal2022graphnli,yin2023annobert,agarwal2024decentralised}.
However, modelling the context dependencies and information flows inherent in conversation trees is a challenging task. Moreover, many of these approaches are based on pre-trained language models (PLMs) such as transformers, where the encodings ignore the distinctive dependency of a comment or reply on another post~\cite{gu-gift-2023}.

In online conversations, posts are responses to other posts and therefore may contain references to, or assume implicit context drawn from them. Intuitively, leveraging the context of the surrounding conversation when fine-tuning PLMs may yield better contextualised representations of conversations. However, existing PLMs such as BERT~\cite{Devlin-bert2019} are designed to handle sequential texts~\cite{gu-gift-2023} but need to be enhanced to encode conversation tree semantics.

Unfortunately, choosing the `right' context is itself a difficult task – choosing the wrong context may lead to noise, while on the other hand ignoring relevant posts could lead to wrong conclusions. In addition, the `right' or appropriate context may differ from one conversation understanding task to the other. In this work, we ask the Research Question (RQ): \textit{Can we effectively capture the conversational context and develop a flexible \textbf{general-purpose mechanism} to learn the right context for \textbf{different online conversation understanding tasks}?}

To learn what aspects of context are important for different kinds of downstream post disambiguation tasks, we propose the notion of \textit{Conversation Kernels}: flexible structures that identify, given a particular post and a particular conversation understanding task, which other posts in the conversation provide the `right' context.
We design two \textit{families} of conversation kernels. The first is built on the concept of node neighborhoods and considers all nodes in the one-hop and two-hop neighborhoods of a post as potential context; the second considers the \textit{tree structure} of online conversations and uses the siblings (posts that share the same parent), children (posts that have replied to the post being categorized), and the ancestral lineage (the parent post which the post being categorized has replied to, its grand parent and so on).
In both cases, the conversation kernel architecture consists of (1) a context retriever module that captures the context through either of the defined kernel shapes, and (2) a transformer-based context-augmented encoder module that maps comments to their contextual embeddings.

To validate our framework, we crawl a \texttt{slashdot.org} corpus of $1954$ conversations, covering the period $2014-2022$ and containing $509,669$ comments in total. We chose \texttt{slashdot.org} as the comments on that site can have multiple different labels applied to them, such as `funny' or `insightful'. Thus, we can train conversation kernels to recognise these very different kinds of comments and thereby explore  the generality and flexibility of the Conversation Kernel framework.

We test our framework's performance on the downstream tasks of learning four different kinds of comments, chosen as exemplars to showcase the generality of the architecture: `funny', `informative', `insightful', and `interesting'. Normally, recognizing vastly different kinds of comments such as `funny' and `informative' might be considered as different NLP tasks that might require different kinds of models or approaches, the conversation kernel architecture is designed to be flexible such that specialized models for each task can be learned using the same approach, thus greatly streamlining the process of conversation understanding.

Experimental results show that context-augmented conversation kernels can significantly outperform baselines such as BERT, RoBERTa and LSTM, with absolute improvements in accuracy up to $20\%$ and up to $19\%$
for macro-F1 scores across the range of the four exemplar tasks. The model (trained on 2014--22 data from \texttt{slashdot.org}) proves to be robust even when tested on previously unseen data from a different time period (Jan -- Nov 2023).

In recent years,  large language models (LLMs) have emerged as efficient zero- and few-shot learners that could potentially achieve best-in-class performance on new text classification tasks, such as labelling different kinds of comments.  However, our evaluations show that conversation kernels also outperform state-of-the-art LLMs including GPT-3.5 and GPT-4, which further highlights that choosing the right context is a hard problem that may be difficult to solve simply by using much larger models than ours.

We believe that the Conversation Kernel approach, of first \textit{learning which parts of the structure of a conversation are relevant context} for a given conversation understanding task and then \textit{augmenting models with this context as additional input}, holds significant promise. The two families of conversation kernel shapes we consider in this paper, as well as the four exemplar tasks we evaluate it on should be seen as proof-of-concept that this approach yields benefits. We fully expect that future research will develop new families of conversation kernels. To enable this line of work and to enhance reproducibility, we have released the \texttt{slashdot} dataset and the model code for non-commercial research\footnote{ The slashdot dataset and code are available at  \url{https://netsys.surrey.ac.uk/datasets/slashdot}.}.

\section{Related Work}\label{sec:relwork}
The significant growth of users interacting on social media platforms has brought increased research interest in extending computational approaches developed for classifying monologic corpora (e.g. news collections~\cite{Choi2010,Awadallah2012,fan2020} and reviews~\cite{mukherjee-liu-2012,wang-ling-2016,popescu2005,Kushal2003}) to the dialogic domain, in order to make sense of such online conversations. Beginning with efforts to classify harmful (hate) speech through keyword-based~\cite{Davidson2017, waseem2016} and statistical mining methods~\cite{mihaylov2015,Xu2010}, or deep neural architectures applied to annotated datasets~\cite{Mozafari2020, caselli2020,wang-ling-2016}, recent efforts have researched adding real-world~\cite{lin-2022} or commonsense~\cite{basu2021-commonsense} knowledge to transformer-based architectures to improve classification performance. These background context-aware methods have been applied to detecting latent hatred in tweets~\cite{lin-2022} and irony or sarcasm in news headlines and Reddit data~\cite{basu2021-commonsense}. These studies highlight the importance of adding context to tackle the challenges of linguistic nuance and diversity, but also recognise that more sophisticated structures are required to capture the information flow between text and knowledge, especially in cases of domain discrepancy between the two~\cite{lin-2022}.

Developing models for conversational dialogic data brings new challenges, with ill-formed sentence structures, higher language variability~\cite{mehdad2013} and limited-length replies or comments that implicitly refer to other posts within the same conversation. The classification label (whether a post is funny or informative, etc.) may also be apparent only in the context of the conversation~\cite{ghosh-cosyn2023}, requiring consideration of both local, i.e. lexical and structural, and global (dialogue act) contextual features~\cite{allen2014}. A notable effort in this direction is the CoSyn model~\cite{ghosh-cosyn2023} that jointly models a user's personal stance with a Fourier attention method and the conversational context using graph convolution networks, to detect implicit hate in a Twitter conversational dataset.

Initial efforts for \Slashdot conversation analysis looked at developing visual interactive systems for analysing conversations (topic with related authors and comments)~\cite{hoque2014}, topic labelling with phrase entailment~\cite{mehdad2013} and assessing the controversiality of posts by calculating the h-index of the corresponding discusssion~\cite{gomez2008}. Studies have also considered the dialogic nature of \Slashdot conversations by applying Discourse Tree theory~\cite{mann1988} for modelling conversations as a collection of linked monologues to detect disagreement~\cite{allen2014}.

A smaller body of work has looked at identifying funny vs. informative/insightful posts by modelling this as a multi-label prediction task~\cite{qin2019} or applying lexical features (polarity, slang, emoticons etc.) to identify funny posts~\cite{reyes2010}. These works reveal that compared to other `funny' disambiguation settings such as one-line jokes or news headlines, lexical features are less discriminatory in conversational data, with the underlying humour mechanism derived from a discrepancy between two viewpoints in conversations~\cite{reyes2010}, rather than linguistic strategies such as irony or sarcasm or socio-cultural context~\cite{vanroy2020-lt3}. Moreover, the funny and informative categories were found to be quite similar~\cite{reyes2010}. All these existing approaches fail to leverage the structural dependencies between posts/replies, and the contextual representations are also not learnt end-to-end.

\section{Slashdot Dataset}\label{sec:slashdot-dataset}
The Slashdot\footnote{\url{https://slashdot.org}, last accessed 22 Mar 2025.} technology-related online news forum enables users to post articles and comment or respond to other users' posts, resulting in a tree-like dialogue structure~\cite{allen2014}. The user moderation and the formalised reply-to structure between comments enable directed and structured conversations~\cite{allen2014}, providing a valuable source for analysing the dynamics of online discussions and the attitudes and behaviours of online communities.
Slashdot is organised into various discussion topics, with popular categories being ``Technology'' (news related to information technology), ``Science" (scientific discoveries and breakthroughs), ``Devices" (hardware and software news), and ``Entertainment" (movies and celebrity culture).

Crucially, the comments and posts are scored (from 1 to 5) and categorized through a community-driven process, with the following tags: `funny', `informative', `insightful', `interesting', `off-topic', `flamebait', and `troll'. This provides us with a unified platform where multiple conversation understanding tasks can be explored, for instance, how to learn whether a post is `funny' or not, `informative' or not, etc.

\subsection{Data Collection Methodology}
Following identification of the discussion topics to be retrieved, our developed data scraping tool retrieves its HTML content with the Selenium Webdriver running on the Chrome browser. The tool design takes into account different DOM structures of each discussion topic and retrieves the complete data for all the comments in a topic. The retrieved HTML content is parsed using the BeautifulSoup library to extract only the required HTML tags for the topic (i.e., topic name, topic id, content, author, and published data) and for each comment inside the topic (comment ID, parent ID, timestamp, discussion topic and text). The target variable in our dataset is ``category”, which represents the comment category (funny, insightful, etc.). The ``score'' variable indicates the community's rating of the comment, with higher scores indicating that the comment is well received by other users. We focus on crawling large conversations with $100$ or more comments.

\subsection{Dataset Statistics and Analysis}
The collected corpus has data of $509,669$ comments from $1954$ conversations from January 2014 to September 2022. The average number of comments per discussion is $261.15$, with a minimum of $101$ comments and a maximum of $864$ comments in a discussion topic. This suggests that engagement levels vary widely among discussion topics, with some generating higher levels of comments than others. The average number of tokens or words per comment is $99.38$.

\begin{figure}
    \centering
    \includegraphics[width=\columnwidth]{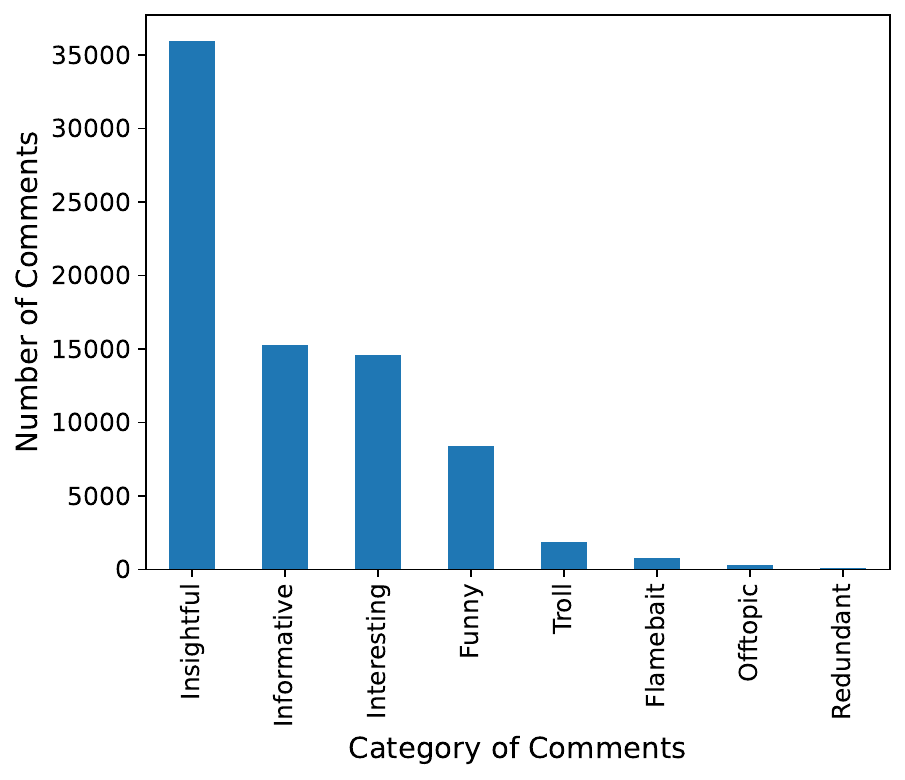}
    \caption{Distribution of comments with respect to different categories.}
    \label{fig:comment-distri-cat}
\end{figure}


Out of $509,669$ comments, only $70,316$ comments have labels based on the nature of the comments. In order to better understand the engagement levels on \Slashdot, we represent the total number of comments corresponding to each category, as shown in Figure~\ref{fig:comment-distri-cat}. The results show that most of the comments fall into the four categories of `insightful', `informative', `interesting', and `funny'. Conversations categorised as `insightful' received the highest number of comments, numbering more than $33,000$. This was followed by `informative' comments at $14,000$, with similar numbers for `interesting', and `funny' comments having the smallest count of the four. In contrast, discussions categorised as `flamebait', `off-topic' and `redundant' account for less than $2\%$ of the total comments, suggesting that \Slashdot users are more likely to engage in discussions that are `informative', `insightful', `interesting', or `funny', and are less likely to engage with discussions that are perceived as being `off-topic' or not relevant.



An analysis of the relationship between score and category in terms of the number of comments shows that posts in the `informative', `insightful', `interesting', and `funny' categories received comments across all scorers with the highest number of comments having a score of 5, followed by a steady decline in the number of comments as the score decreases. The remaining categories, on the other hand, show comment score distribution between 1 and 2, with minimal or no comments with other scores. These patterns suggest that \Slashdot users are more likely to engage with `informative', `insightful', `interesting', and `funny' comments.


\begin{figure}
    \centering
    \begin{subfigure}{\linewidth}
        \centering
        \includegraphics[width=\textwidth]{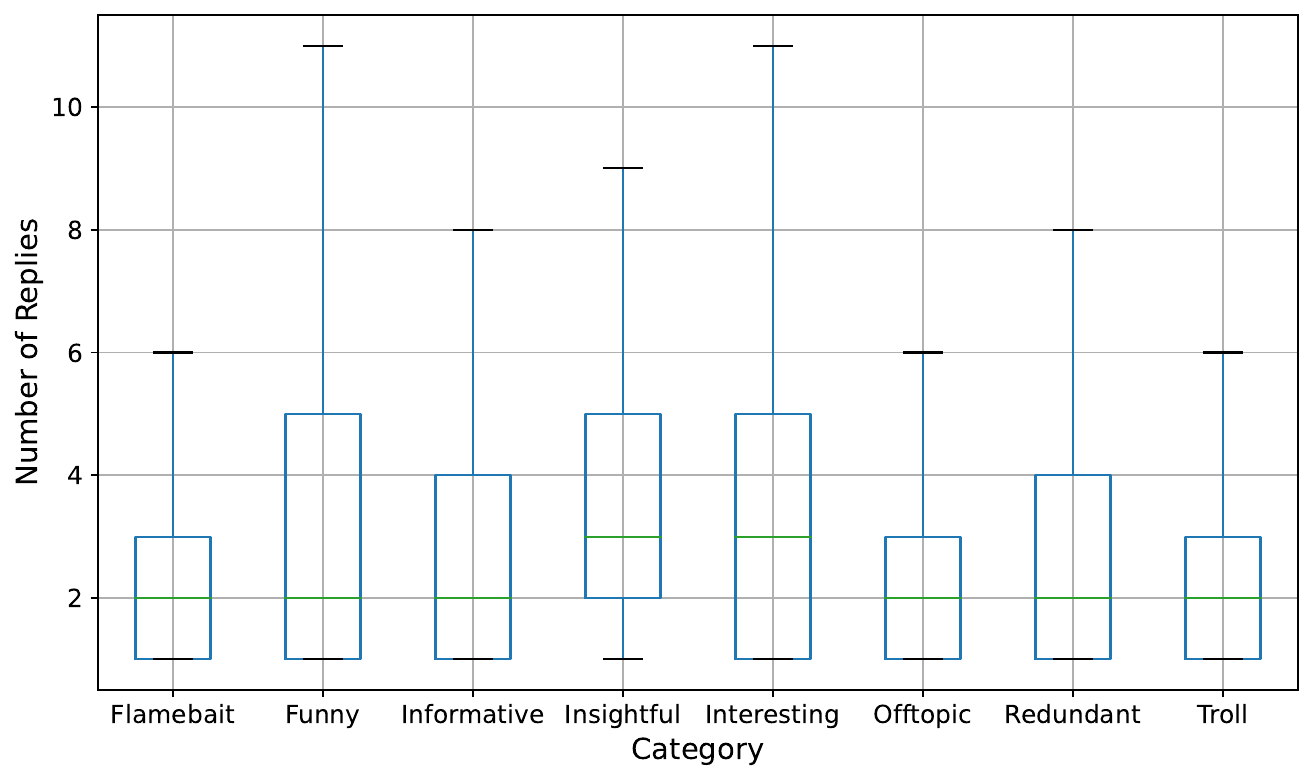}
    \end{subfigure}
    \begin{subfigure}{\linewidth}
        \centering
        \includegraphics[width=\textwidth]{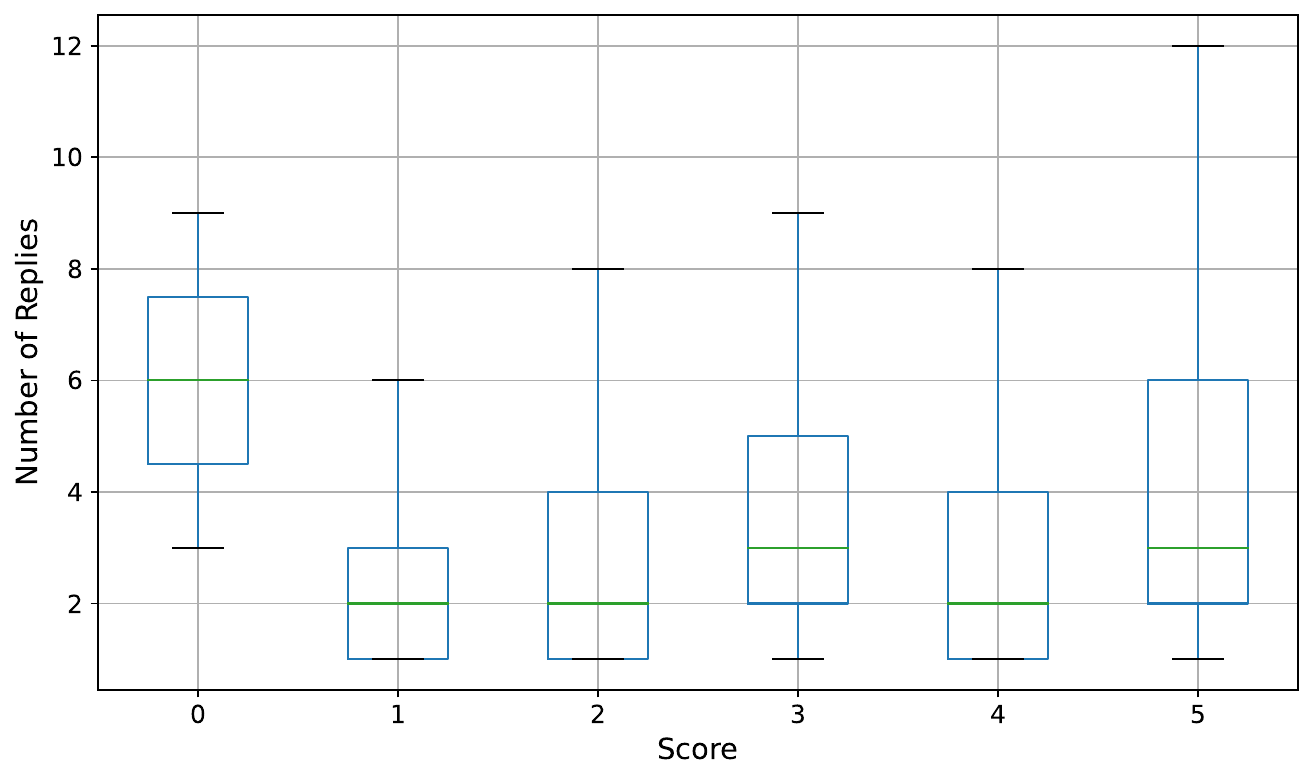}
    \end{subfigure}
    \caption{Box plots showing the number of replies for each category and score.}
    \label{fig:boxplots}
\end{figure}

Figure~\ref{fig:boxplots} provides a visual representation of the distribution of replies within each category and score. The box plots show the median and quartiles of the number of replies, enabling identification of 
categories receiving the highest or the lowest number of replies. For instance, a higher median of the `insightful' category suggests that users are more likely to engage with and reply to insightful comments. Similarly, users are more engaged with comments having scores of 3 and 5.

Following these findings, we concentrate our analysis of \Slashdot conversations to the four major comment categories which attract the highest distribution of comments and replies:
`informative', `insightful', `interesting', and `funny'.

\subsection{Problem Statement}

We frame our problem statement as that of developing a common framework to formulate the context window discovery task concerned with \textit{\textbf{comment nature prediction}} for a diverse set of comments. We instantiate the framework for mining \Slashdot conversational content to determine whether a comment is insightful, interesting, informative, or funny. Predicting the nature of comments is a context-dependent task and requires understanding of the conversational context to be able to predict whether a comment is insightful, interesting, funny, etc. We first pre-process \Slashdot conversations to convert them into \textit{conversation trees} using comment IDs and parent IDs obtained while crawling these conversations. A conversation tree~\cite{agarwal2022graphnli,boschi2021has,agarwal2023graph,agarwal2024gascom} is a tree structure where nodes are the comments and a directed edge from a node to its parent indicates that the node replies to its parent comment. We then input these conversation trees into our framework which we discuss next.

\section{Conversation Kernels}\label{sec:convokernel}
In this section, we introduce the concept of conversation kernels and describe its model architecture. The Conversation Kernel  has 2 components: conversational context retriever (described in Section~\ref{sec:context-retriever}), which extracts the relevant conversational context driven by different kernel shapes; context-augmented encoder (Section~\ref{sec:context-encoder}), which encodes the conversational context together with the target comment for online conversation understanding.

\begin{figure}
    \centering
    \includegraphics[width=\linewidth]{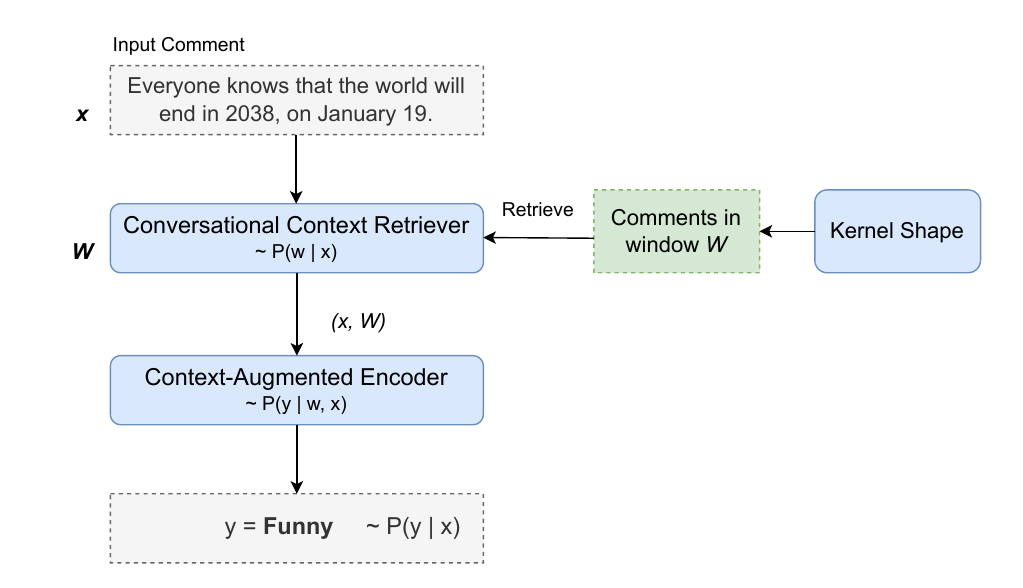}
    \caption{Conversation Kernels}
    \label{fig:conv-kernels}
\end{figure}

The conversation kernel architecture (Fig.~\ref{fig:conv-kernels}) takes a target comment $x$ as input and learns a probability distribution $p(y | x)$ over all possible values of $y$. In our task of comment nature prediction, $y$ is a binary variable (whether a comment is funny or not, insightful or not, and so on). This decomposes the computation of $p(y | x)$ into two steps: \textit{retrieval} followed by \textit{encoder} to predict the nature of comments in online conversations. The conversational context retriever module uses different kernel shapes to choose a set of windows or shapes to capture the conversational context. Let $W$ be the set of $n$ windows: $W = \{w_1, w_2, ..., w_n\}$ and each window has fixed number of comments $L$: $w_i = \{c_1, c_2, ..., c_L\}$. Given a target comment $x$, we first retrieve relevant conversational context windows $w$ from the set $W$. We model this as a sample from the distribution $p(w | x)$. Then, we condition on both the conversational context $w$ and the target comment $x$ to predict the output $y$ as $p(y | x, w)$. To obtain the overall likelihood of predicting $y$, we treat $w$ as a latent variable and marginalize over all the possible values of $w$ as per below:

\begin{equation} \label{eq1}
    p(y | x) = \sum_{w \in W}^{} p(y | w, x) p(w | x)
\end{equation}

\subsection{Conversational Context Retriever}\label{sec:context-retriever}

The conversational context retriever module models $p(w | x)$. To capture the relevant conversational context, the retriever module uses two different kinds of \textit{\textbf{kernel shapes}} as follows:

\begin{figure}
    \centering
    \includegraphics[width=\linewidth]{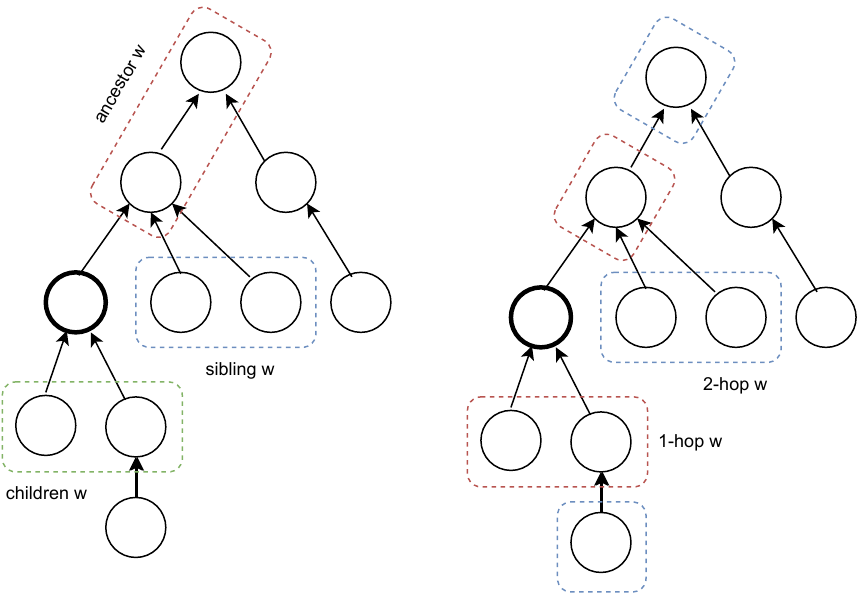}
    \caption{Illustration of different kernel shapes. Different windows are depicted by different colors. Left: ancestor (red), sibling (blue), children (green) windows; Right: one-hop (red), two-hop (blue) windows. Target comment node is in bold.}
    \label{fig:kernel-shape}
\end{figure}

\subsubsection{\textbf{Ancestors, siblings \& children windows}} This kernel shape has 3 windows each for the ancestors, siblings, and children nodes in a conversation tree, as shown in Figure~\ref{fig:kernel-shape} (left). Each window $w$ contains at most $L$ comments from a conversation tree. If a window has less than $L$ comments, it chooses all of them. But if a window has more than $L$ comments, it chooses the first $L$ comments based on the timestamp. It is important to fix the window size $L$ since online conversations can grow up to hundreds and thousands of comments (nodes in a conversation tree). \textit{Ancestor} window chooses $L$ ancestors of the target node starting from its parent in a conversation tree. \textit{Sibling} window chooses $L$ sibling nodes for the target node except itself. \textit{Children} window chooses $L$ children of the target node based on the timestamp. In case the target node is a leaf node, no nodes will be chosen by the \textit{children} window.

\subsubsection{\textbf{One- \& two-hop neighborhood windows}} This kernel shape has 2 windows for each of the one-hop and two-hop neighbors of the target node in a conversation tree, as shown in Figure~\ref{fig:kernel-shape} (right). Again, each window $w$ contains $L$ comments. \textit{One-hop} window selects first $L$ direct neighbors of the target node based on the timestamp in a conversation tree. Similarly, \textit{two-hop} window selects first $L$ two-hop neighbors of the target node.

We propose these two kinds of kernel shapes because they capture neighboring conversational context differently. For example, ancestor-sibling-children windows are capable of capturing far away ancestor and children nodes that are even three or more hops away. On the other hand, one- and two-hop windows capture neighborhood nodes that are local (one or two hops away) to the target node. We envision that new kernel shapes can be developed in the future based on different online conversation understanding tasks as different kernel shapes may be helpful for different kinds of tasks.

Overall, the retriever module is defined using a dense inner product model once it captures the context through either of the kernel shapes as shown below.

\begin{equation} \label{eq2}
    p(w | x) = \mathit{softmax}(f(x, w))
\end{equation}

\begin{equation} \label{eq3}
    f(x, w) = \mathit{Embed}_{\mathit{comment}}(x)^T \mathit{Embed}_{\mathit{window}}(w)
\end{equation}

In equation~\ref{eq3}, $\mathit{Embed}_{\mathit{comment}}$ and $\mathit{Embed}_{\mathit{window}}$ are embedding functions mapping the target comment and comments in a window $w$ to fix-sized vectors. The relevance score $f(x, w)$ is the inner product of vector embeddings of $x$ and $w$. This relevance score assigns different weights to different kinds of nodes based on the conversational context and the task. The retrieval distribution $p(w | x)$ is the softmax over all relevance scores with respect to each of the windows $w \in W$, as in equation~\ref{eq2}.

For embedding functions, we use the transformer-based RoBERTa base model~\cite{liu2019roberta} to map comments to their corresponding contextual embeddings. We use the RoBERTa model to generate contextual embeddings because it outperforms transformer-based BERT and LSTM models (see Table~\ref{tab:results}). We then take embeddings corresponding to the $[CLS]$ token denoted as $\mathit{RoBERTa}_{\mathit{[CLS]}}$. Finally, we perform a linear projection of the output embeddings, denoted as a projection matrix \textbf{W} as shown in equation~\ref{eq4}.

\begin{equation} \label{eq4}
    \mathit{Embed}_{\mathit{comment}}(x) = \textbf{W}_{\mathit{comment}} \mathit{RoBERTa}_{\mathit{[CLS]}}(x)
\end{equation}

To compute embedding for a window $w$ denoted by $Embed_{window}$, we have $L$ comments. Once we get $[CLS]$ token embeddings from the RoBERTa model for each of the comments $c_i^w \in w$, we take a mean of their embeddings to get a resultant embedding because it works better than max pooling in our experiments. Again, this resultant embedding is linearly projected using a projection matrix \textbf{W} as shown:
$\mathit{Embed}_{\mathit{window}}(w) = \textbf{W}_{\mathit{window}} \mathit{Mean}_{c_i^w \in w} ( \mathit{RoBERTa}_{\mathit{[CLS]}}(c_i^w))$.


\subsection{Context-Augmented Encoder}\label{sec:context-encoder}

Given a target comment $x$ and a retrieved context window $w$, the context-augmented encoder models $p(y | w, x)$. It also uses the transformer-based RoBERTa~\cite{liu2019roberta} model for mapping comments to their contextual embeddings. Firstly, it concatenates the target comment $x$ with comments $c_i^w$ where $i \in [1, L]$ in a context window $w$ separated by $[SEP]$ tokens as shown below:
\begin{equation} \label{eq6}
    join_{\mathit{RoBERTa}}(x, w) = [CLS] x [SEP] c_1^w [SEP] ... c_L^w [SEP]
\end{equation}
Then this concatenated text is input into the RoBERTa model. The resultant embeddings corresponding to the $[CLS]$ token are extracted and assigned as shown below:

\begin{equation} \label{eq7}
\begin{split}
    \mathit{Embed}_{\mathit{encoder}}(x, w) = \mathit{RoBERTa}_{\mathit{[CLS]}}(\\
    \mathit{join}_{\mathit{RoBERTa}}(x, w) )
\end{split}
\end{equation}

Finally, these contextual embeddings are input into a fully-connected layer, followed by softmax for predicting the probabilities of output variable $y$:
\begin{equation} \label{eq8}
    p(y | w, x) = \mathit{softmax}( \mathit{MLP}( \mathit{Embed}_{\mathit{encoder}}(x, w) ) )
\end{equation}

\section{Experiments and Results}\label{sec:eval}

\subsection{Experimental Setup}\label{sec:exp-setup}

Firstly, we split the conversation trees from \Slashdot into 80:10:10 split for training, validation, and testing sets, respectively. We treat the context window discovery task for each category as a binary classification problem with an appropriate balanced dataset, e.g., for context discovery of `funny' comments, we randomly select equal numbers of 'funny' and non-funny (i.e. sampling from the other categories) comments to make a balanced dataset for model input. We then input conversation trees from the training set into the conversation kernels model and train both the retriever and the encoder modules together in an end-to-end fashion. We use a batch size of $16$, Adam optimizer with learning rate $1 \times 10^{-5}$, window size $L=3$ and a linear learning rate warm-up over $10\%$ of the training data. We experiment with different values of $L$ ranging from 2 to 10 and find that $L=3$ is performing the best. We experiment with two different kinds of kernel shapes as discussed in Section~\ref{sec:context-retriever}. We make our model end-to-end trainable by minimizing the binary cross-entropy loss computed based on the model predictions and the ground-truth labels. We implement the model using Transformers~\cite{wolf-etal-2020-transformers} and PyTorch~\cite{Paszke_PyTorch_An_Imperative_2019} libraries and train it for $3$ epochs. We use NVIDIA Titan RTX GPU with 24 GB of memory for training.

\subsection{Baselines and Evaluation Metrics}
We compare our conversation kernels with the following relevant baselines:

\textit{LSTM}~\cite{lstm1997}: The Long Short Term Memory (LSTM) model is effective for multi-class classification tasks.
The input text is pre-processed to remove stop words and the maximum length of the text sequences after tokenization is set to 256, with an embedding dimension of 100. The target labels are one-hot encoded. The model is trained with the Adam optimiser and mean squared error as the loss function.

\textit{BERT}~\cite{Devlin-bert2019}: The pre-trained Bidirectional Encoder Representations from Transformers (BERT) is the state-of-the-art model for sequence classification tasks.
We input individual \Slashdot comments, setting the maximum sequence length to 75 and use Adam optimiser with cross-entropy as the loss function.

\textit{RoBERTa}~\cite{liu2019roberta}: RoBERTa is a modified version of BERT model, giving state-of-the-art performance in various classification tasks. Similar to BERT, we input individual \Slashdot comments, setting the maximum sequence length to 75 and use Adam optimiser with cross-entropy as the loss function.

\textit{RoBERTa + context}: This uses RoBERTa, but with additional conversational context of the parent comment. The input to the model is a comment and its parent separated by the \texttt{[SEP]} token.

\textbf{Evaluation metrics}: We compare our conversation kernels method to the above baselines in terms of classification accuracy and the macro-F1 score. The macro-F1 score is reported as a single score that balances both precision and recall metrics and because it treats each class equally, regardless of its frequency or imbalance in the dataset.

\subsection{Results}\label{sec:results}

\begin{table*}[]
    \centering
    \small
    \begin{tabular}{l|cc|cc|cc|cc}
        \hline
          & \multicolumn{2}{|c}{\textbf{Insightful}} & \multicolumn{2}{|c}{\textbf{Informative}} & \multicolumn{2}{|c}{\textbf{Interesting}} & \multicolumn{2}{|c}{\textbf{Funny}}  \\
        \hline
         \textbf{Model} & \textbf{Acc.} & \textbf{macro-F1} & \textbf{Acc.} & \textbf{macro-F1} & \textbf{Acc.} & \textbf{macro-F1} & \textbf{Acc.} & \textbf{macro-F1}    \\
        \hline
         LSTM & 0.5590 & 0.5518 & 0.5988 & 0.6218 & 0.5780 & 0.5814 & 0.7406 & 0.7368 \\
         BERT & 0.6345 & 0.6219 & 0.6997 & 0.6996 & 0.6320 & 0.6318 & 0.7665 & 0.7553 \\
         RoBERTa & 0.6351 & 0.6278 & 0.7059 & 0.7058 & 0.6437 & 0.6403 & 0.7691 & 0.7610  \\
         RoBERTa + context & 0.6361 & 0.6191 & 0.6965 & 0.6958 & 0.6366 & 0.6366 & 0.7698 & 0.7682  \\
        \hline
         CK: anc-sib-child windows & \textbf{0.6481} & \textbf{0.6330} & \textbf{0.7896} & \textbf{0.7804} & \textbf{0.7607} & \textbf{0.7520} & 0.7742 & 0.7741 \\
         CK: 1-hop 2-hop windows & 0.6319 & 0.6320 & 0.7211 & 0.7005 & 0.6713 & 0.6461 & \textbf{0.7957} & \textbf{0.7954} \\
           \hline
         GPT-3.5 (post + full conversation) & 0.5293 & 0.5292 & 0.6227 & 0.6144 & 0.5360 & 0.4985 & 0.7747 & 0.7742 \\
         GPT-4 (post + full conversation) & 0.5520 & 0.5328 & 0.6220 & 0.6185 & 0.5620 & 0.4993 & 0.7933 & 0.7895 \\
        \hline

    \end{tabular}
    \caption{Accuracy and macro-F1 scores for different comment categories. CK denotes Conversation Kernels. The GPT-3.5 and GPT-4 results are based on a random 10\% stratified sample of the entire dataset.}
    \label{tab:results}
\end{table*}

Table~\ref{tab:results} compares the performance of the conversation kernels with the baselines. Among the baseline models, RoBERTa performs the best in terms of macro-F1 scores for `insightful', `informative', and `interesting' categories. For `funny' category, RoBERTa with additional context performs the best. Our proposed conversation kernel outperforms all the baselines both in terms of accuracy and macro-F1 score, showcasing its effectiveness in modeling the conversational context for comment nature prediction, for all the four conversation categories. We experiment with conversation kernels of two different kinds of kernel shapes as discussed in Section~\ref{sec:context-retriever}.

The results show an improvement of $11.17\%$ for the `interesting', and $7.46\%$ for the `informative' category, for macro-F1 scores against the best performing RoBERTa baseline, validating its ability to minimise both the false positive and false negative rate. The model also slightly outperforms the baseline RoBERTa model in detecting insightful comments. In the case of `funny' comments, our conversation kernel model shows an impressive performance on both the accuracy ($0.7957$) and macro-F1 ($0.7954$) scores.

It is interesting that the ancestor-child-sibling windows are the best performing family of kernel shapes for interesting, informative and insightful categories, whereas the local neighborhoods of the comment (1-hop and 2-hop windows) are a more useful discriminative feature for distinguishing funny comments from non-funny ones. Therefore, different kinds of kernel shapes may be useful for different kinds of conversation understanding tasks.

\begin{figure}
    \centering
    \includegraphics[width=\linewidth]{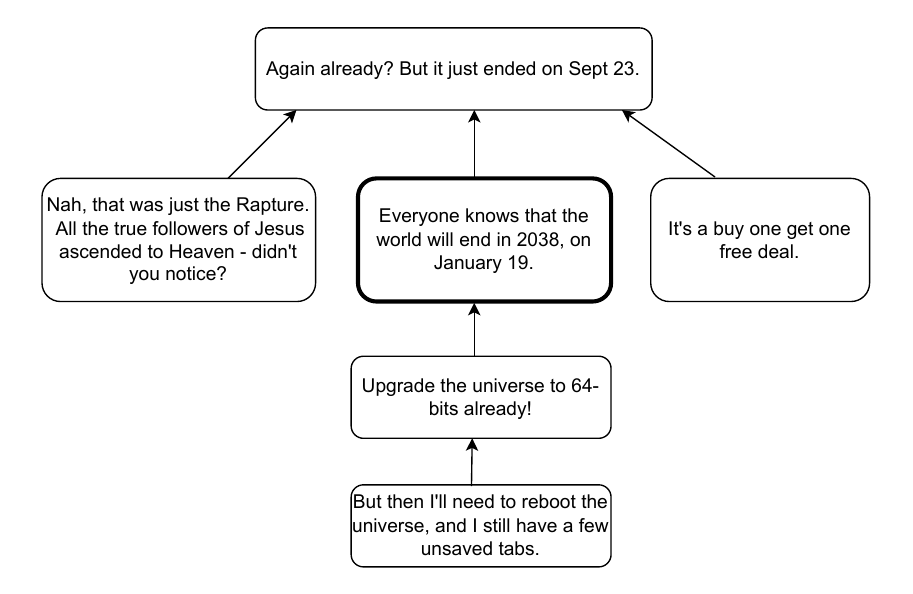}
    \caption{An example `funny' Slashdot conversation.}
    \label{fig:eg-funny-conv}
\end{figure}

To understand why, we highlight one example `funny' comment (in bold border) in Figure~\ref{fig:eg-funny-conv}. The original post (parent of the post being considered) has posted a URL of a website announcing the end of the world, and our bolded post has posted a funny reply. Notice that not only the comment being considered, but also all the other comments in the two hop neighborhood are funny, tongue-in-cheek comments responding back to the original post, or to the bolded post we are looking at. Given this common pattern that one funny post attracts other funny responses, the local one- and two-hop neighborhood performs better for `funny' comments. It can also be seen that each funny comment is relatively self-contained, and can be understood without too much additional context; thus the more local one- and two-hop neighborhoods perform well.

\subsection{Generalizability of Conversation Kernels}\label{sec:generalizability}

To show generalizability, we crawl an additional latest snapshot of Slashdot data from January to November 2023 containing $13,962$ comments, as a sample from another time period. We find that our conversation kernel, trained on the \Slashdot dataset from 2014 to 2022, performs just as well on the latest snapshot of the data which is previously unseen. Detailed performance results for conversation kernel with ancestor-sibling-child windows are shown in Table~\ref{tab:generalizability-results}. Other social media platforms such as Reddit, X (Twitter), etc. follow a similar tree structure of online conversations wherein a comment may attract multiple replies but it can reply to exactly one parent comment leading to a multi-threaded tree structure. Therefore, our conversation kernels would also generalize to these social media platforms, enabling us to understand online conversations.

\begin{table*}[]
    \centering
    \small
    \begin{tabular}{l|cc|cc|cc|cc}
        \hline
          & \multicolumn{2}{|c}{\textbf{Insightful}} & \multicolumn{2}{|c}{\textbf{Informative}} & \multicolumn{2}{|c}{\textbf{Interesting}} & \multicolumn{2}{|c}{\textbf{Funny}}  \\
        \hline
         \textbf{Dataset} & \textbf{Acc.} & \textbf{macro-F1} & \textbf{Acc.} & \textbf{macro-F1} & \textbf{Acc.} & \textbf{macro-F1} & \textbf{Acc.} & \textbf{macro-F1}    \\
        \hline
         Jan-Nov 2023 & 0.6418 & 0.6310 & 0.7818 & 0.7784 & 0.7597 & 0.7509 & 0.7734 & 0.7730  \\
        \hline
    \end{tabular}
    \caption{Performance of conversation kernels (with ancestor-sibling-child windows) on the latest Slashdot dataset from January to November 2023.}
    \label{tab:generalizability-results}
\end{table*}

\begin{figure}[t]
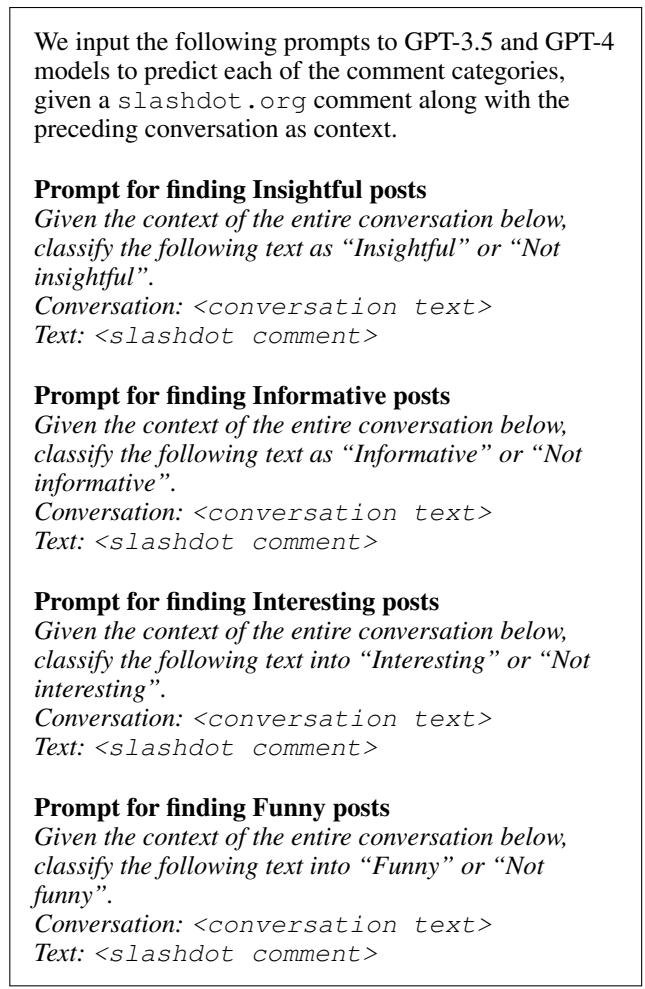

    \raggedright
    \begin{framed}
We input the following prompts to GPT-3.5 and GPT-4 models to predict each of the comment categories, given a \Slashdot comment along with the preceding conversation as context.
\bigskip

\textbf{Prompt for finding Insightful posts}

\textit{Given the context of the entire conversation below, classify the following text as ``Insightful'' or ``Not insightful''.}

\textit{Conversation: \texttt{<conversation text>}}

\textit{Text: \texttt{<slashdot comment>}}

\bigskip
\textbf{Prompt for finding Informative posts}

\textit{Given the context of the entire conversation below, classify the following text as ``Informative'' or ``Not informative''.}

\textit{Conversation: \texttt{<conversation text>}}

\textit{Text: \texttt{<slashdot comment>}}

\bigskip
\textbf{Prompt for finding Interesting posts}

\textit{Given the context of the entire conversation below, classify the following text into ``Interesting'' or ``Not interesting''.}

\textit{Conversation: \texttt{<conversation text>}}

\textit{Text: \texttt{<slashdot comment>}}

\bigskip
\textbf{Prompt for finding Funny posts}

\textit{Given the context of the entire conversation below, classify the following text into ``Funny'' or ``Not funny''.}

\textit{Conversation: \texttt{<conversation text>}}

\textit{Text: \texttt{<slashdot comment>}}
\end{framed}
    \caption{Prompts for comment nature prediction tasks.}
    \label{fig:prompts}
\end{figure}

\subsection{Comparison with LLMs}
Given our goal of a general-purpose mechanism for discovering context relevant to different tasks, it is natural to ask whether general-purpose pre-trained Large Language Models (LLMs)~\cite{naveed2023comprehensive,brown2020language}, which have been proven to excel at a wide variety of tasks~\cite{zhu2023can,agarwal2023haterephrase,agarwal2024medhalu,agarwal2024codemirage}, could discover the right conversation context. To test this, we perform a further baseline comparison, asking GPT-3.5~\cite{ouyang2022training} and GPT-4~\cite{achiam2023gpt} to predict the comment nature. Using the prompts in Figure~\ref{fig:prompts}, we provide these models with the comment together with including the entire conversation as possible context for LLMs.

To keep costs down, we performed this test on a random $10\%$ sample of the entire dataset. LLMs also have a limit of $8192$ tokens; thus we are not able to provide the entire conversation as input for long conversations. This affected $81$ conversations. For these conversations, we first linearize the entire conversation tree by ordering comments in temporal order. To predict the comment nature for a given post, we provide as context as many immediately preceding comments of the post as would fit into the LLM token limit.

Table~\ref{tab:results} shows that although GPT-4 consistently performs better than GPT-3.5 model as expected, both LLMs do not reach the performance obtained by  Conversation Kernels even though the LLMs have sight of the entire conversation. LLM performance for identifying `funny' posts is roughly similar to Conversation Kernels,  which could be explained by the fact that  funny posts are usually self-contained and can be understood as funny without reference to surrounding posts for context. However, for the other three categories (`insightful', `informative' and `interesting'), Conversation Kernels offer a $10-15\%$ higher accuracy and macro-F1 scores, indicating the benefit of learning the `right' conversation context.

\section{Conclusions}\label{sec:conc}

This paper presents a unified approach to the problem of conversation understanding, by developing a two-step methodology of first understanding relevant conversation context relevant to a task and then utilizing that context in the downstream task. We propose Conversation Kernels as principled and generalizable kernel shapes that are useful in picking up as context all relevant comments surrounding a particular post in a conversation thread that we are interested in. Conversation kernel shapes are designed to first retrieve comments that are ``close by'' (\textit{i.e.,} in the neighbourhood)  the post of interest, and then an attention mechanism is used to give additional weight to those that are more relevant. We show how this  can be applied as a uniform approach to train models for detecting widely different kinds of comments such as `informative', `insightful', `interesting' or `funny'.

To circumvent the problem that many conversation classes may be difficult to define precisely, we build, as our first contribution, a unique dataset of over 70,000 \Slashdot posts, with examples of what Slashdot users considered to be `informative', `insightful', `interesting' and `funny'. To enable reproducibility and further research, we share this dataset at  \texttt{\url{https://netsys.surrey.ac.uk/datasets/slashdot}}.

Although there are eight different labels that users can apply to comments on \Slashdot (including labels such as `troll' or `flamebait'), our exploratory characterization reveals that users mostly engage with posts from four categories: `informative', `insightful', `interesting' and `funny'. As such, we set the task of developing a machine learning (ML) pipeline that can predict whether or not a post is considered to fall into one of these four categories by users on \Slashdot.

The key contribution of the Conversation Kernel architecture is the development of a generalizable approach for detecting relevant context needed for deeper conversation understanding when posts often refer to other posts --- for example, a reply may only be funny in the context of the post it is replying to. As proof of concept of the efficacy of our approach, we develop two families of \textit{kernel shapes} to retrieve comments surrounding a post that is being classified, and perform the classification of a post after augmenting it with context built up from the retrieved surrounding concept.  The first kernel shape we develop uses ancestors, siblings and children nodes of a post as context windows. The second uses one-hop and two-hop neighborhoods of the post in question.

Our evaluation shows that conversation kernels outperform other relevant baselines such as LSTM, BERT and RoBERTa with additional context for all categories we consider. Ancestor, sibling and children context windows perform the best for categorizing posts as insightful, informative and interesting, whereas the 1-hop and 2-hop neighborhood windows perform the best for funny posts. We also show that the Conversation Kernel approach outperforms much larger LLMs, showcasing the difficulty and importance of retrieving the \textit{right} context.

We believe that the conversation kernels approach introduced in this paper is generalizable in two ways: First, the two families of kernel shapes we introduced in this work are merely intended as proof-of-concepts. We aim to explore other kernel window shapes, including strategies of mixing and matching windows across different families, as well as exploring the relevance of comments from non-local windows. Second, we believe that the conversation kernel approach can be applied to other subjective labels as well as more objective topics. We will demonstrate this by adapting our method to other datasets and also to other important and well studied tasks in conversation understanding, such as identifying spam, misinformation and hate speech, especially in cases where hate or misinformation may be `implicit' and conveyed with reference to the parent or other nearby posts. Also, we compare conversation kernels with LLMs in a zero-shot setting using prompting. However, fine-tuning of the LLMs is also possible using techniques such as Low-rank adaptation (LoRA)~\cite{hu2021lora} and we would like to explore interesting possibilities of integrating conversation kernels with LLMs and explore related directions for selecting appropriate conversational context.

\subsection{Limitations}\label{sec:limitations}


In forums such as BBC’s \textit{Have Your Say?}\footnote{\url{https://www.bbc.co.uk/blogs/haveyoursay/archives.html}, last accessed 22 Mar 2025.}, there is no explicit threaded reply structure, requiring us to infer from the text of a reply which other post it is replying to, to construct conversation trees. In this less restrictive user interface, a single post may refer to or reply to multiple other posts, creating more than one edge and a conversation structure that is no longer a tree but a more general graph. We believe that conversation kernels would work in this more general context as well, with 1-hop and 2-hop windows sampling all the available conversational context. However, this has not been tested empirically.

Conversation kernels make use of conversation context from surrounding posts. While conversation kernels can label each post as a conversation evolves and new posts are added, it becomes more effective only after a reasonable number of replies have been added. At the beginning of a conversation, when not a lot of conversational context is available, conversation kernels will likely to perform similar to baseline models such as RoBERTa, which also operate without the additional context.

Currently, conversation kernels are trained on English conversations. However, with widespread multilingual online conversations and conversations in low resource languages, there is a need to build models for online conversation understanding in multilingual and low resource settings. It is easy to adapt conversation kernels for low resource and multilingual settings by using language models trained on specific languages in the retriever and encoder components of the model.




\bibliography{aaai25}

\section*{Ethical Statement}
Although we believe that Conversation Kernels are an abstract approach which then need to be applied in different application contexts where sensitive issues of ethics and legality might apply, we conclude by considering two such issues for completeness.

Firstly, this paper intentionally applies the Conversation Kernel approach to relatively uncontroversial kinds of comments such as `funny' or `informative'. However, as highlighted above, the approach is generalizable to other tasks such as detecting hate speech, including in highly sensitive contexts such as political conversations~\cite{pushkalHateSpeech} where greater care will need to be taken to ensure that the \textit{right} context is considered, as mistakes of both omission (not detecting a hate speech act) and commission (wrongly detecting a valid or legal post as hate speech) can have disastrous consequences. This requires further empirical examination and is beyond the scope of the current paper.

Secondly, as with many other AI/ML models, the efficacy and correctness of conversation kernels greatly depends on the underlying data used to train the model. Thus, the training dataset and its biases need to be kept in mind for any downstream applications. For example, what is considered `funny' by Slashdot users (who are mostly from the tech community) may not align with other communities.

Despite the above `obvious' limitations and ethics considerations, we believe the generalizability of the conversation kernel approach, as well as its efficacy in a wide variety of conversation classifications, makes it a useful addition to the arsenal of tools being developed for online conversation understanding.

\end{document}